\documentclass[journal]{IEEEtran}
\usepackage{amsmath,graphicx,amsfonts,epsfig,epstopdf,datetime,algorithm,algorithmic,multirow,xcolor,subfig}

\begin{document}
%
\title{VMF-SNE:  Embedding for Spherical Data}
%
%
%

\author{Mian~Wang,
        Dong~Wang~\IEEEmembership{Member,~IEEE}
\thanks{Mian Wang is a master student at Beijing University of Posts and Telecommunications.
This work was done when he was a visiting student at Tsinghua University. E-mail: wm@cslt.riit.tsinghua.edu.cn}
\thanks{Dong Wang is with the center for speech and language technology (CSLT),  Tsinghua University. E-mail: wangdong99@mailst.singhua.edu.cn}
\thanks{Manuscript received --; revised --}}


\maketitle

\begin{abstract}
T-SNE is a well-known approach to embedding high-dimensional data and has been widely used in data visualization. The basic assumption of t-SNE is that the data are non-constrained in the Euclidean space and the local proximity can be modelled by Gaussian distributions. This assumption does not hold for a wide range of data types in practical applications, for instance spherical data for which the local proximity is better modelled by
the von Mises-Fisher (vMF) distribution instead of the Gaussian. This paper presents a vMF-SNE embedding algorithm to embed spherical data. An iterative process is derived to produce an efficient embedding. The results on a simulation data set demonstrated that vMF-SNE produces better embeddings than t-SNE for spherical data.

\end{abstract}

\begin{IEEEkeywords}
data embedding, data visualization, t-SNE, Von Mises-Fisher distribution
\end{IEEEkeywords}

\IEEEpeerreviewmaketitle

\section{Introduction}

\IEEEPARstart{H}{igh-dimensional}
data embedding is a challenging task in machine learning and is important for many applications,  particularly data visualization. Principally, data embedding involves projecting high-dimensional data to a low-dimensional (often 2 or 3) space where the major structure (distribution) of the data in the original space is mostly preserved. Therefore data embedding can be regarded as a special task of dimension reduction, with the objective function set to preserve the structure of the data.

Various traditional dimension reduction approaches can be used to perform data embedding, e.g., the principal component analysis (PCA)~\cite{hotelling1933analysis} and the multi-dimensional scaling (MDS)~\cite{borg2005modern}. PCA finds low-dimensional embeddings that preserve the data covariance as much as possible. Classical MDS finds embeddings that preserve inter-sample distances, which is equivalent to PCA if the distance is Euclidean. Both the PCA and MDS are simple to implement and efficient in computation, and are guaranteed to discover the true structure of data lying on or near a linear subspace. The shortage is that they are ineffective for data within non-linear manifolds.

A multitude of non-linear embedding approaches have been proposed. The first approach is to derive the global non-linear structure from local proximity. For example, ISOMAP extends MDS by calculating similarities of distant pairs based on similarities of neighbouring pairs~\cite{tenenbaum1998mapping,tenenbaum2000global}. The self-organizing map (SOM) or Kohonen net extends PCA and derives the global non-linearity by simply ignoring distant pairs~\cite{kohonen1982}. The same idea triggers the generative topographic mapping (GTM)~\cite{bishop1998gtm}, where the embedding problem is cast to a Bayesian  inference with an EM procedure. The local linear embedding (LLE) follows the same idea but formulates the embedding as a local-structure learning based on linear prediction~\cite{roweis2000nonlinear}. Another approach to deriving the global non-linear structure involves various kernel learning methods, e.g., the semi-definite embedding based on kernel PCA~\cite{weinberger2004learning} and the colored maximum variance unfolding (CMVU)~\cite{song2007colored}.

A major problem of the above non-linear embedding methods is that most of them are not formulated in a probabilistic way, which leads to potential problems in generalizability. The stochastic neighbor embedding (SNE)~\cite{hinton2002stochastic} attempts to solve the problem.
It models local proximity (neighbourhood) of data in both the original and embedding space by Gaussian distributions, and the embedding process minimizes the kullback-leibler (KL) divergence of the distributions in the original space and the embedding space.

A potential drawback of SNE is the `crowding problem', i.e., the data samples tend to be crowded together in the embedding space~\cite{van2008visualizing}. A UNI-SNE approach was proposed to deal with the problem, which introduces a symmetric cost function and a smooth model when computing similarities between the images (embeddings) of data in the embedding space~\cite{cook2007visualizing}. With the same problem in concern, \cite{van2008visualizing} proposed t-SNE, which also uses a symmetric cost function, but employs a Student t-distribution rather than a Gaussian distribution
to model similarities between images. T-SNE has shown clear superiority over other embedding
methods particularly for data that lie within several different but related low-dimensional manifolds.

Although highly effective in general, t-SNE is weak in embedding data that are not Gaussian. For example,  there are many applications where the data are distributed on a hyper-sphere, such as the topic vectors in document processing~\cite{reisinger2010spherical} and the normalized i-vectors in speaker recognition~\cite{dehak2011front}.
These spherical data are naturally modelled by the von Mises Fisher (vMF) distribution rather than the Gaussian~\cite{fisher1995statistical,dhillon2003modeling,mardia2009directional}, and hence are unsuitable to be embedded
by t-SNE. This paper presents a vMF-SNE algorithm to embed spherical data. Specifically, the Gaussian distribution and the Student t-distribution used by t-SNE in the original and the embedding space respectively are all replaced by vMF distributions, and an EM-based optimization process is derived to conduct the embedding. The experimental results on simulation data show that vMF-SNE produces better embeddings for spherical data. The code is  online available\footnote{http://cslt.riit.tsinghua.edu.cn/resources.php?Public\%20tools}.

The rest of the paper is organized as follows. Section~\ref{sec:rel} describes the related work, and Section~\ref{sec:theory} presents the vMF-SNE algorithm. The experiment is presented in~\ref{sec:exp}, and the paper is concluded in Section~\ref{sec:con}.

\section{Related work}
\label{sec:rel}

This work belongs to the extensively studied area of dimension reduction and data embedding. Most of the related work in this field has been mentioned in the last section. Particularly, our work is motivated by t-SNE~\cite{van2008visualizing}, and is designed specifically to embed spherical data which are not suitable to be processed by t-SNE. A more related work is the parametric embedding (PE)~\cite{iwata2007parametric}, which embeds vectors of posterior probabilities, thus sharing  a similar goal as our proposal: both attempt to embed data in a constrained space though the constrains are different ($\ell$-1 in PE and $\ell$-2 in vMF-SNE).


Probably the most relevant work is the spherical semantic embedding (SSE)~\cite{le2014semantic}. In the SSE approach, document vectors and topic vectors
are constrained on a unit sphere and are assumed to follow the vMF distribution.  The topic model and the embedding model are then jointly optimized in a
generative model framework by maximum likelihood. However, SSE infers  local similarities
between data samples (document vectors in~\cite{le2014semantic})
using a pre-defined latent structure (topic vectors), which is difficult to be generalized to other tasks as the
latent structure in most scenarios is not available. Additionally,
the cost function of SSE is the likelihood, while vMF-SNE uses the symmetric KL divergence.

\section{vMF-distributed stochastic neighbouring embedding}
\label{sec:theory}

\subsection{t-SNE and its limitation}

Let $\{x_i\}$ denote the data set in the high-dimensional space, and $\{y_i\}$ denote the corresponding embeddings, or images. The t-SNE algorithm
measures the pairwise similarities in the high-dimension space as the joint distribution of $x_i$ and $x_j$ which is assumed to be Gaussian, formulated by the following:

\begin{equation}
\label{eq:p}
p_{ij} = \frac{e^{-||x_i-x_j||^2/{2\sigma^2}}}{\sum_{m \neq n}{e^{-||x_m-x_n||^2/{2\sigma^2}}}}.
\end{equation}

\noindent In the embedding space, the joint probability of $y_i$ and $y_j$ is modelled by a Student t-distribution with one degree of freedom, given by:

\begin{equation}
\label{eq:q}
q_{ij} = \frac{(1+||y_i-y_j||^2)^{-1}}{\sum_{m \neq n}{(1+||y_m-y_n||^2)^{-1}}}.
\end{equation}

\noindent The cost function of the embedding is the KL divergence between $p_{i,j}$ and $q_{i,j}$, which is formulated by:

\[
KL(P||Q) = \sum_i \sum_j p_{ij} ln \frac{p_{ij}}{q_{ij}}.
\]

\noindent A gradient descendant approach has been devised to conduct the optimization, which is fairly efficient~\cite{van2008visualizing}. Additionally, the symmetric form of Eq. (\ref{eq:p}) and the long-tail property of the Student t-distribution alleviate the crowding problem suffering the original SNE and other embedding approaches.

The assumption that t-SNE holds deserves highlight: the joint probabilities of the original data and the embeddings follow a Gaussian distribution and a Student t-distribution, respectively. This is generally fine in most scenarios, however for data that are confined in a non-linear subspace, this assumption is potentially invalid and the t-SNE embedding is no longer optimal. This paper focuses on spherical data embedding, for which t-SNE tends to fail. This is because the Gaussian distribution assumed by t-SNE can hardly model spherical data, and the Euclidean  distance associated with Gaussian distributions is not appropriate to measure similarities on a hyper-sphere. A new embedding algorithm is proposed, which shares the same embedding framework as t-SNE, but uses a more appropriate distribution form and a more suitable similarity measure to model spherical data.

\subsection{vMF-SNE}

It has been shown that the vMF distribution is a better choice than the Gaussian in modelling spherical data, and the associated cosine distance is better than the Euclidean distance when measuring similarities in a hyper-spherical space, for instance, in tasks such as spherical data clustering~\cite{strehl2000impact,banerjee2005clustering}. Therefore, we present an embedding method based on the assumption that the data in both the original and the embedding space follow vMF distributions. This new method is thus called `vMF-SNE'.

Mathematically, the probability density function of the vMF distribution on the ($d$-1)-dimensional sphere in $R^d$ is given by:

\[
  f_d(x;\mu,\kappa) = C_d(\kappa)e^{\kappa \mu^T x}
\]
\noindent where $||x||=||\mu||=1$, $\kappa > 0$ and $\mu$ are parameters of the distribution and $C_d(\kappa)$ is a normalization constant. Note that the vMF distribution implies the cosine
distance. As in t-SNE, the symmetric distance is used in both the original and embedding space. In the original space, define the conditional probability of $x_j$ given $x_i$ as:

\begin{equation}
\label{eq:vmfpp}
p_{j|i} = \frac{f_d(x_j;x_i,\kappa_i)}{\sum_{m \neq i} f_d(x_m; x_i, \kappa_i)},
\end{equation}
\noindent the joint distribution $p_{ij}$ is defined as follows:

\begin{equation}
\label{eq:vmfp}
  p_{ij}=\frac{p_{i|j}+p_{j|i}}{2}.
\end{equation}

\noindent In the embedding space, a simpler form of joint distribution is chosen by setting the concentration parameter $k_i$ the same for all $y_i$. This choice follows t-SNE, and the rationale is that the distribution $p_{j|i}$ in the original space needs to be adjusted according to the data scattering around $x_i$. However, doing so in the embedding space will cause unaffordable complexity in computation, as we will see shortly. The joint distribution $q_{ij}$
with this simplification is given by:

\begin{equation}
\label{eq:vmfq}
  q_{ij}=\frac{e^{\kappa y_i^T y_j}}{\sum_{m{\neq}n}e^{\kappa y_m^T y_n}}.
\end{equation}

As in t-SNE, the KL divergence between the two distributions is used as the cost function:
\[
  \mathcal{L}=\sum_i\sum_j p_{ij}ln\frac{p_{ij}}{q_{ij}}
\]
\noindent By gradient descendant, minimizing $\mathcal{L}$ with respect to $\{y_i\}$ leads to the optimal embedding.
The gradients will be derived in the following section.

\subsection{Gradient derivation}

First note that

\[
\mathcal{L}=\sum_{i,j} p_{ij} ln (p_{ij}) - \sum_{i,j}p_{ij}ln(q_{ij}).
\]
\noindent Since the first item on the right hand side of the equation is in dependent of the embedding, minimizing $\mathcal{L}$ equals to
maximizing the following cost function:

\[
\mathcal{\tilde{L}} = \sum_{i,j}p_{ij}ln(q_{ij}).
\]

\noindent Define $Z=\sum_{m{\neq}n}e^{ky_m^Ty_n}$, we have:

\[
\mathcal{\tilde{L}} =\kappa \sum_{i,j}p_{ij}  y_i^T y_j - ln Z,
\]

\noindent where $\sum_{i,j}p_{ij}=1$ has been employed. The gradient of $\mathcal{\tilde{L}}$ with respect to the embedding $y_k$ is then derived as:

\begin{eqnarray}
\frac{\partial{\mathcal{\tilde{L}}}}{\partial{y_k}} &=& 2 \kappa \sum_{i} p_{ik}  y_i - \frac{1}{Z} \frac{\partial{ln Z}}{\partial{y_k}} \\
                                                   &=& 2 \kappa \sum_{i}  p_{ik}  y_i - \frac{2\kappa }{Z} \{\sum_{i} e^{\kappa y_i^T y_k}  y_i \} \\
                                                   \label{eq:g}
                                                   &=& 2 \kappa \sum_{i} (p_{ik}-q_{ik})  y_i
\end{eqnarray}
\noindent This is a rather simple form and the computation is efficient. Note that this simplicity is partly due to the identical $\kappa$ in the embedding space, otherwise the computation will be very demanding.

Algorithm~\ref{alg:vmf} illustrates the vMF-SNE process. Notice that in the original data space, $\kappa_i$ is required. Following~\cite{van2008visualizing},
$\kappa_i$ is set to a value that makes the perplexity $\mathcal{P}_i$  equal to a pre-defined value $\mathcal{P}$, where $\mathcal{P}_i$ is formulated by:

\begin{equation}
\label{eq:ppl}
\mathcal{P}_i = 2^{H(p_{j|i})}
\end{equation}
\noindent and $H(\cdot)$ is the information entropy defined by:
\[
H(p_{j|i})={-\sum_j p_{j|i} log_2 (p_{j|i}})
\]
\noindent where $p_{j|i}$ has been defined in Eq. (\ref{eq:vmfpp}). As mentioned in~\cite{van2008visualizing}, making the perplexity associated to each data point the same value normalizes the data scattering and so benefits outliers and crowding areas.

\begin{algorithm}[h]
    \caption{vMF-SNE}
    \label{alg:vmf}
    \begin{algorithmic}[1]
        \REQUIRE ~~
            \\
            Input:\\
            $\{x_i; ||x_i||=1, i=1,...,N\}$: data to embed\\
            $\mathcal{P}$: perplexity in the original space\\
            $\kappa$: concentration parameter in the embedding space\\
            T: number of iterations \\
            $\eta$: learning rate\\
            Output:\\
            $\{y_i; ||y_i||=1, i=1,...,N\}$: data embeddings
        \ENSURE ~~
        \STATE compute $\{\kappa_i\}$ according to Eq. (\ref{eq:ppl})
        \STATE compute $p_{ij}$ according to Eq. (\ref{eq:vmfp}), and set $p_{ii}=0$
        \STATE randomly initialize $\{y_i\}$

        \FOR{$t=1$ to T}
                \STATE compute $q_{ij}$ according to Eq. (\ref{eq:vmfq})
                \FOR{$i=1$ to N}
                    \STATE $\delta_{i}=\frac{\partial{\mathcal{\tilde{L}}}}{\partial{y_i}}$ according to Eq. (\ref{eq:g})
                    \STATE $y_i=y_i + \eta \delta_{i}$
                 \ENDFOR
        \ENDFOR
    \end{algorithmic}
\end{algorithm}

\section{Experiment}
\label{sec:exp}

To evaluate the proposed method, we employ vMF-SNE to visualize spherical data and compare it with the traditional t-SNE.  Since visualization is not a quantitative evaluation, an entropy-based criterion is proposed to compare the two embedding approaches.

\subsection{Simulation data}

The experiments are based on simulation data. The basic idea is to sample $k$ clusters of data and examine if the cluster structure can be preserved after embedding. The sampling process starts from the centers of the $k$ clusters, i.e., $\{\mu_i; ||\mu_i||=1, i=1,...,k\}$. Although the sampling for different $\mu_i$ is essentially independent, we adopt a different approach: firstly sample the first center $\mu_1$, and then derive other centers $\{\mu_i\}$ by randomly selecting a subset of the dimensions of $\mu_1$ and flipping the signs of the values on these dimensions. By this way, the centers $\{\mu_i\}$ are ensured to be separated on the hyper-sphere, which generates a clear cluster structure associated with the data.

Once the cluster centers are generated, it is easy to sample the data points for each cluster following the vMF distribution. A toolkit provided by Arindam Banerjee and Suvrit Sra was adopted to conduct the vMF sampling\footnote{http://suvrit.de/work/soft/movmf}.
In this work, the dimension of the data is set to $50$, and $800$ data points are sampled in total. The concentration parameter $\kappa$ used in the sampling also varies, in order to investigate the performance of the embedding approaches in different overlapping conditions.

\subsection{Visualization test}

\begin{figure}[!t]
\centering
\includegraphics[width=1.4in]{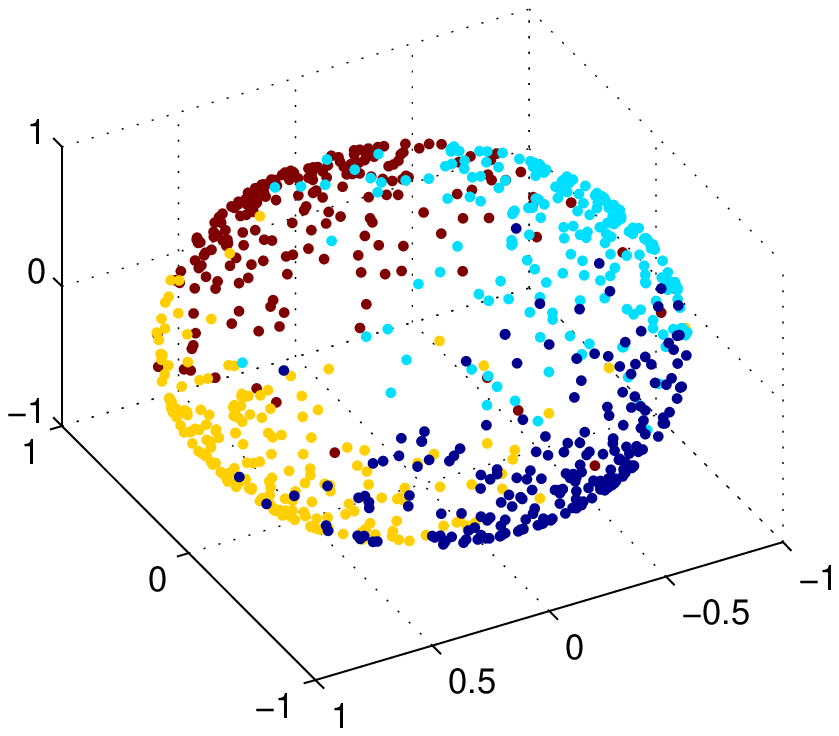}
\hspace{0.5in}
\includegraphics[width=1.4in]{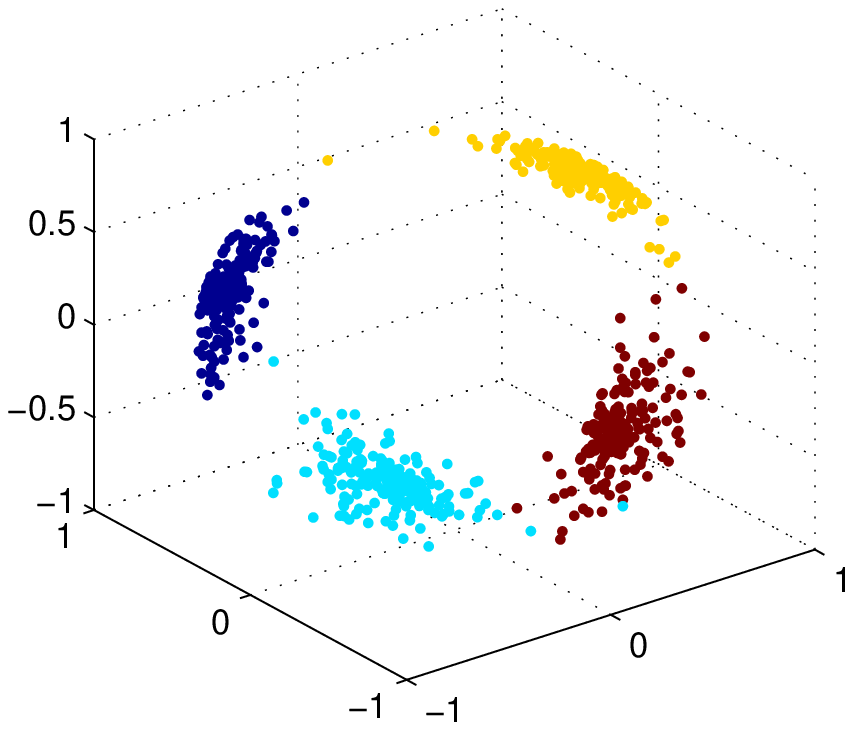}
\caption{The 3-dimensional embedding with vMF-SNE, with data generated following a vMF distribution by setting $\kappa=15$ (left) and $\kappa=40$ (right). The original dimension is $50$, and there are $4$ clusters, each of which is represented by a particular color.}
\label{fig:vmf}
\end{figure}

The first experiment visualizes the spherical data with vMF-SNE. The perplexity $\mathcal{P}$ is set to $40$, and the value of $\kappa$ in the embedding space is fixed to $2$ (see Algorithm~\ref{alg:vmf}). The data are generated following vMF distributions by setting the scattering parameter $\kappa$ to different values. Fig.~\ref{fig:vmf} presents the embedding results on 3-dimensional spheres with vMF-SNE, where the two pictures show the results with $\kappa$=15 and $\kappa$=40 respectively. Note that the $\kappa$ here is used in data sampling, neither the $\kappa$ used to model the original data (which is computed from $\mathcal{P}$ for each data point) nor the $\kappa$ used to model the embedding data (which has been fixed to $2$). It can be seen that vMF-SNE indeed preserves the cluster structure of the data in the embedding space, and not surprisingly, data generated with a larger $\kappa$ are more separated in the embedding space.

\begin{figure}[!t]
\centering
\includegraphics[width=1.4in]{fig/k15vmfsne.eps}
\hspace{0.5in}
\includegraphics[width=1.4in]{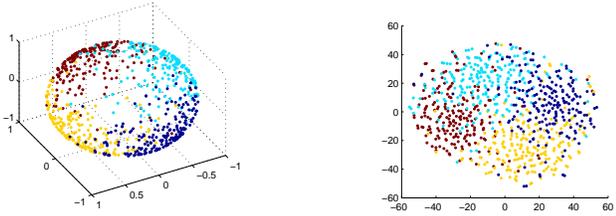}
\caption{The 3-dimensional embedding with vMF-SNE (left) and 2-dimensional embedding with t-SNE (right). The data was generated following a vMF distribution by setting $\kappa=15$.}
\label{fig:vmf-tsne1}
\end{figure}

\begin{figure}[!t]
\centering
\includegraphics[width=1.4in]{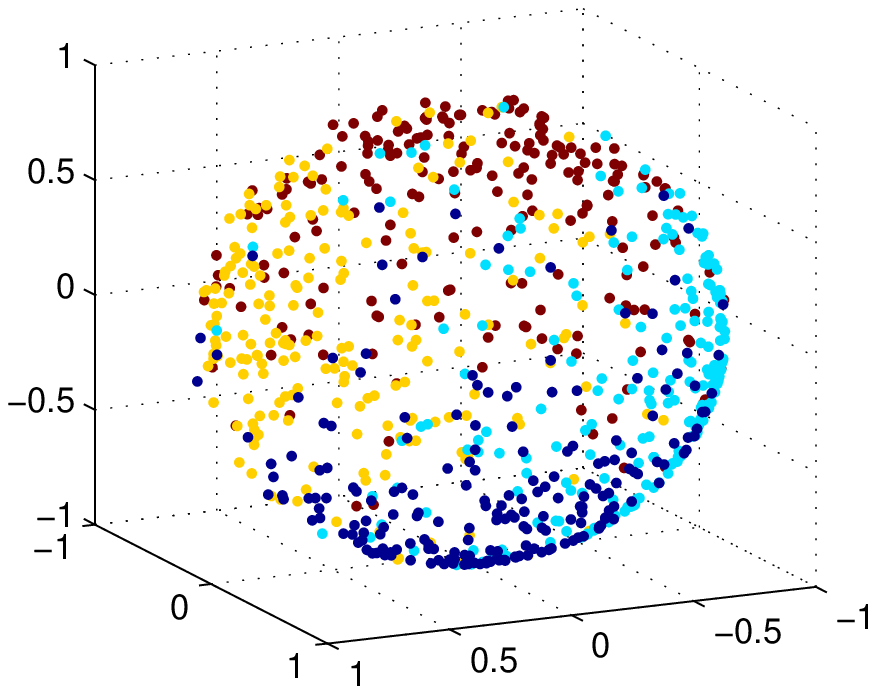}
\hspace{0.5in}
\includegraphics[width=1.4in]{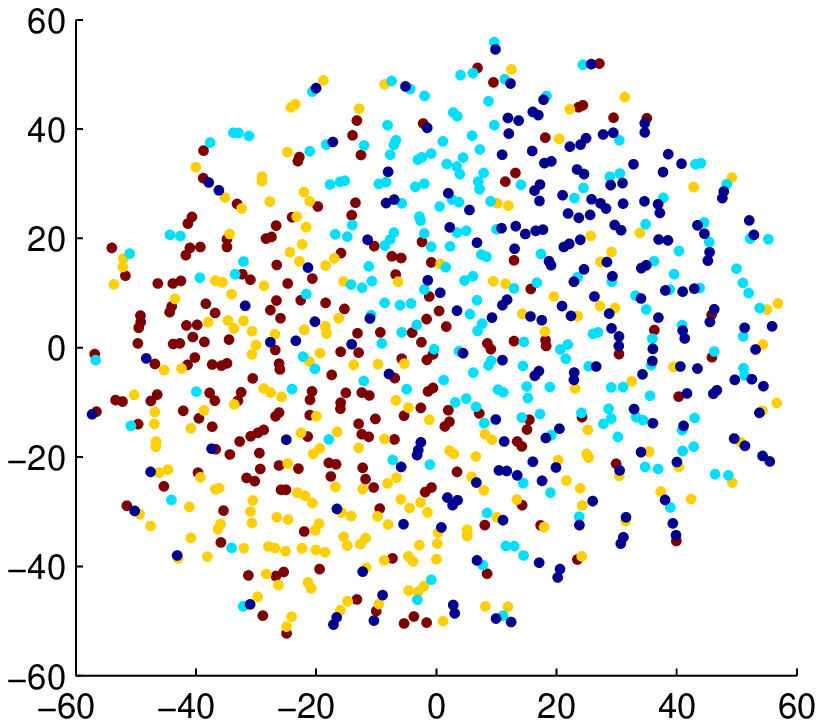}
\caption{The 3-dimensional embedding with vMF-SNE (left) and 2-dimensional embedding with t-SNE (right). The data was generated following a vMF distribution by setting $\kappa=10$.}
\label{fig:vmf-tsne2}
\end{figure}

For comparison, the same data are embedded with t-SNE in 2-dimensional space.  The tool provided by Laurens van der Maaten is used to conduct the embdding\footnote{http://lvdmaaten.github.io/tsne/}, where the perplexity is set to $40$.  The comparative results are shown in Fig.~\ref{fig:vmf-tsne1} and Fig.~\ref{fig:vmf-tsne2} for data generated by setting $\kappa$=15 and $\kappa$=10 respectively. It can be observed that when $\kappa$ is large (Fig.~\ref{fig:vmf-tsne1}), both vMF-SNE and t-SNE perform well and the cluster structure is clearly preserved. However when $\kappa$ is small (Fig.~\ref{fig:vmf-tsne2}), vMF-SNE shows clear superiority. This suggests that t-SNE is capable to model spherical data if the structure is clear, even if the underling distribution is non-Gaussian; however in the case where the structure is less discernable in the high-dimensional space, t-SNE tends to mess the boundary while vMF-SNE still works well.

\subsection{Entropy and accuracy test}

Visualization test is not quantitative. For further investigation, we propose to use the clustering accuracy and entropy as the criteria to measure the quality of the embedding. This is achieved by first finding the images of the
cluster centers, and then classifying the data according to their distances to the centers in the embedding space. The classification accuracy is computed as the proportion of the data that are correctly classified. The entropy of the $i$-th cluster is computed as $H(i) = \sum_{j=1}^{k} c(i,j) ln (c(i,j))$ where $c(i,j)$ is the proportion of the data points generated from the $j$-th cluster but are classified as the $i$-th cluster in the embedding space. The entropy of the entire data set is computed as the average of $H(i)$ over all the clusters.  Table~\ref{tab:vmf-cluster} presents the results. It can be observed that in the case of $4$ clusters, vMF-SNE achieves lower entropy and better accuracy than t-SNE when $\kappa$ is small. If $\kappa$ is large, both the two methods can achieve good performance, for the reason that we have discussed.

In the case of $16$ clusters, it is observed that vMF-SNE outperforms t-SNE with small $\kappa$ values (large overlaps). This seems an interesting property and demonstrates that using the matched distribution (vMF) is helpful to improve embedding for overlapped data. However, with $\kappa$ increases, vMF-SNE can not reach a performance as good as that obtained by t-SNE. A possible reason is that the large number of clusters leads to data crowding which can be better addressed with  the long-tail Student t-distribution used by t-SNE. Nevertheless, this requires further investigation.

\begin{table}[!t]
\center
\caption{Results of Entropy and Accuracy}
\label{tab:vmf-cluster}
\begin{tabular}{|l|l|l|l|l|}
  \hline
  4 Clusters & \multicolumn{2}{|c|}{Entropy} & \multicolumn{2}{|c|}{Accuracy}\\
  \hline
  $\kappa$ & t-SNE & vMF-SNE & t-SNE & vMF-SNE\\
  \hline
    10 &  0.6556  & 0.5922 & 42\% & 64.13\% \\
    20 &  0.4725  & 0.4187 & 85.38\% & 92.63\% \\
    30 &  0.3804  & 0.3676 & 97.38\% & 98.5\% \\
    40 &  0.3485  &  0.3466 & 99.75\% & 99.95\% \\
  \hline
  16 Clusters & \multicolumn{2}{|c|}{Entropy} & \multicolumn{2}{|c|}{Accuracy}\\
  \hline
  10  &   0.3152 & 0.2975 & 15.5\% & 16.88\% \\
  20  &   0.2812 & 0.2608 & 38.25\%  & 40.75\% \\
  30  &   0.2312 & 0.2383 & 68.25\% & 55.13\% \\
  40  &   0.1964 & 0.2187 & 91.25\% & 60.63\% \\
  \hline
\end{tabular}
\end{table}

\section{Conclusions}
\label{sec:con}

A vMF-SNE algorithm has been proposed for embedding high-dimensional spherical data. Compared with
the widely used t-SNE, vMF-SNE assumes vMF distributions and cosine similarities with the
original data and the embeddings, hence suitable for spherical data embedding.
The experiments on a simulation data set demonstrated that the proposed approach works
fairly well. Future work involves studying long-tail vMF distributions to handle crowding data,
as t-SNE does with the Student t-distribution.



\ifCLASSOPTIONcaptionsoff
  \newpage
\fi

\newpage



%

\bibliographystyle{IEEEtran}
\bibliography{mybib}

\end{document}